\documentclass[10pt,twocolumn,letterpaper]{article}
\usepackage{cuted}
\usepackage{cvpr}      
\usepackage{float}
\usepackage[dvipsnames]{xcolor}

\usepackage{caption}
\usepackage{graphicx}
\usepackage{caption}
\usepackage{mathrsfs}
\usepackage{multirow}
\usepackage{pifont}
\usepackage{booktabs}
\usepackage{tabularx}
\usepackage{makecell}
\usepackage[table]{xcolor}
\usepackage{bm}
\usepackage{soul}
\definecolor{cvprblue}{rgb}{0.21,0.49,0.74}
\usepackage[pagebackref,breaklinks,colorlinks,citecolor=cvprblue]{hyperref}

\def\paperID{171} 
\def\confName{3DV\xspace}
\def\confYear{2026\xspace}

\title{Self-Supervised Implicit Attention Priors for Point Cloud Reconstruction}

\author{Kyle Fogarty\\
University of Cambridge\\
\and
Chenyue Cai\\
Princeton University\\
\and 
Jing Yang\\
University of Cambridge\\
\and
Zhilin Guo\\
University of Cambridge\\
\and
Cengiz Öztireli\\
University of Cambridge\\}

\begin{document}


\twocolumn[{%
\renewcommand\twocolumn[1][]{#1}%
\maketitle
\begin{center}
  \includegraphics[width=0.83\textwidth]{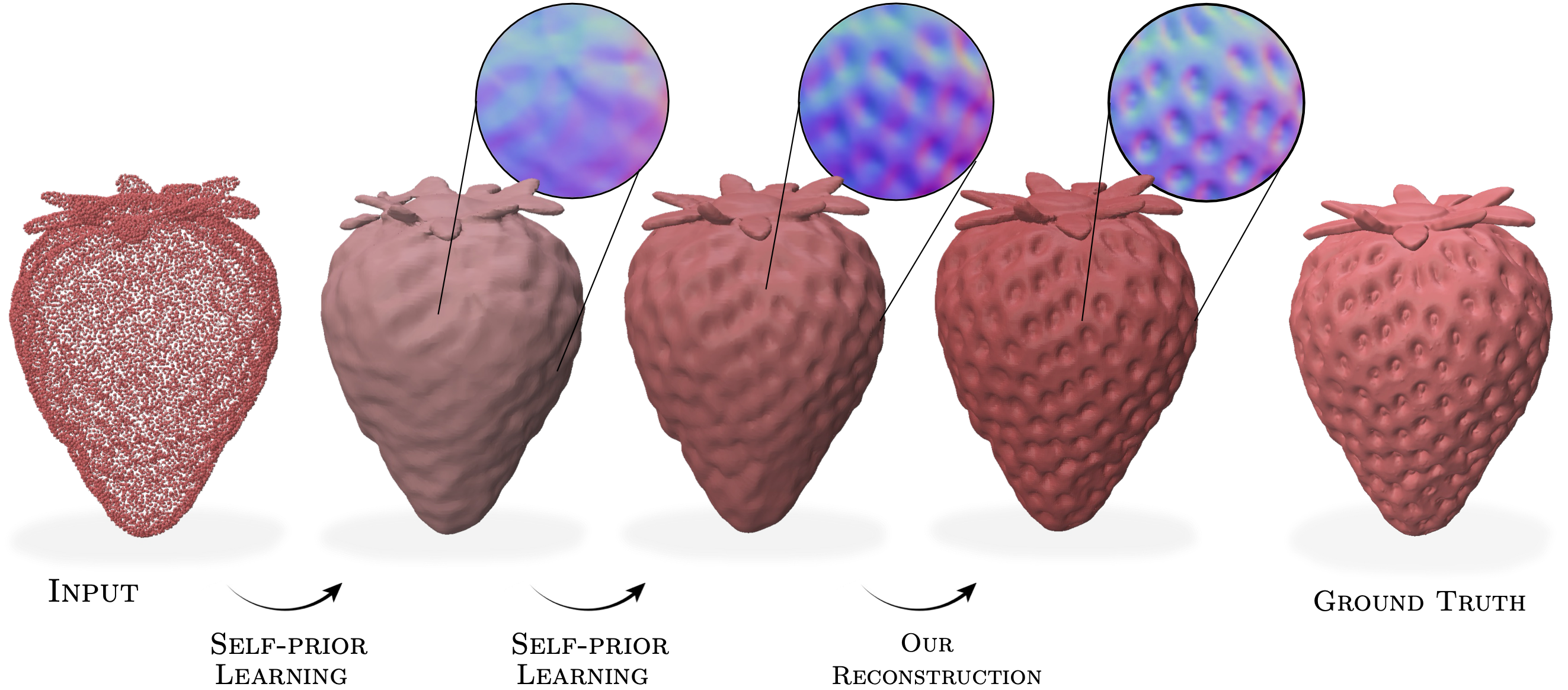}%
  \captionsetup{type=figure}%
  \caption{We propose a novel approach that incorporates a shape-specific self-prior for reconstructing high-fidelity surfaces from irregular point clouds. It \textit{iteratively} leverages shape-specific priors via cross-attention with a compact, learnable dictionary, capturing repeating structures without external training data.}%
\end{center}
\label{fig:teaser}
}]
\begin{abstract}
 {Recovering high-quality surfaces from irregular point cloud is ill-posed unless strong geometric priors are available. We introduce an implicit \textit{self-prior} approach that distills a shape-specific prior directly from the input point cloud itself and embeds it within an \textit{implicit} neural representation. This is achieved by jointly training a small dictionary of learnable embeddings with an implicit distance field; at every query location, the field attends to the dictionary via cross-attention, enabling the network to capture and reuse repeating structures and long-range correlations inherent to the shape. Optimized solely with self-supervised point cloud reconstruction losses, our approach requires no external training data. 
 To effectively integrate this learned prior while preserving input fidelity, the trained field is then sampled to extract densely distributed points and analytic normals via automatic differentiation. We integrate the resulting dense point cloud and corresponding normals into a robust implicit moving least squares (RIMLS) formulation. We show this hybrid strategy preserves fine geometric details in the input data, while leveraging the learned prior to regularize sparse regions. 
 Experiments show that our method outperforms both classical and learning-based approaches in generating high-fidelity surfaces with superior detail preservation and robustness to common data degradations.}

\end{abstract}

\section{Introduction}
\label{sec:introduction}
Recovering continuous surface geometry from discrete, unstructured point clouds is a fundamental problem in computer graphics and 3D vision, crucial for numerous downstream applications. However, the inherent challenges of noise, sparsity, and outliers render this task ill-posed. Consequently, successful reconstruction algorithms must incorporate a \textit{prior} to regularize the problem and guide the surface reconstruction process \cite{berger2014state}. These priors represent assumptions about the underlying geometry, influencing the trade-offs between fidelity to the input data and the plausibility of the resulting surface.\\

\begin{figure}
    \centering
    \includegraphics[width=0.85\linewidth]{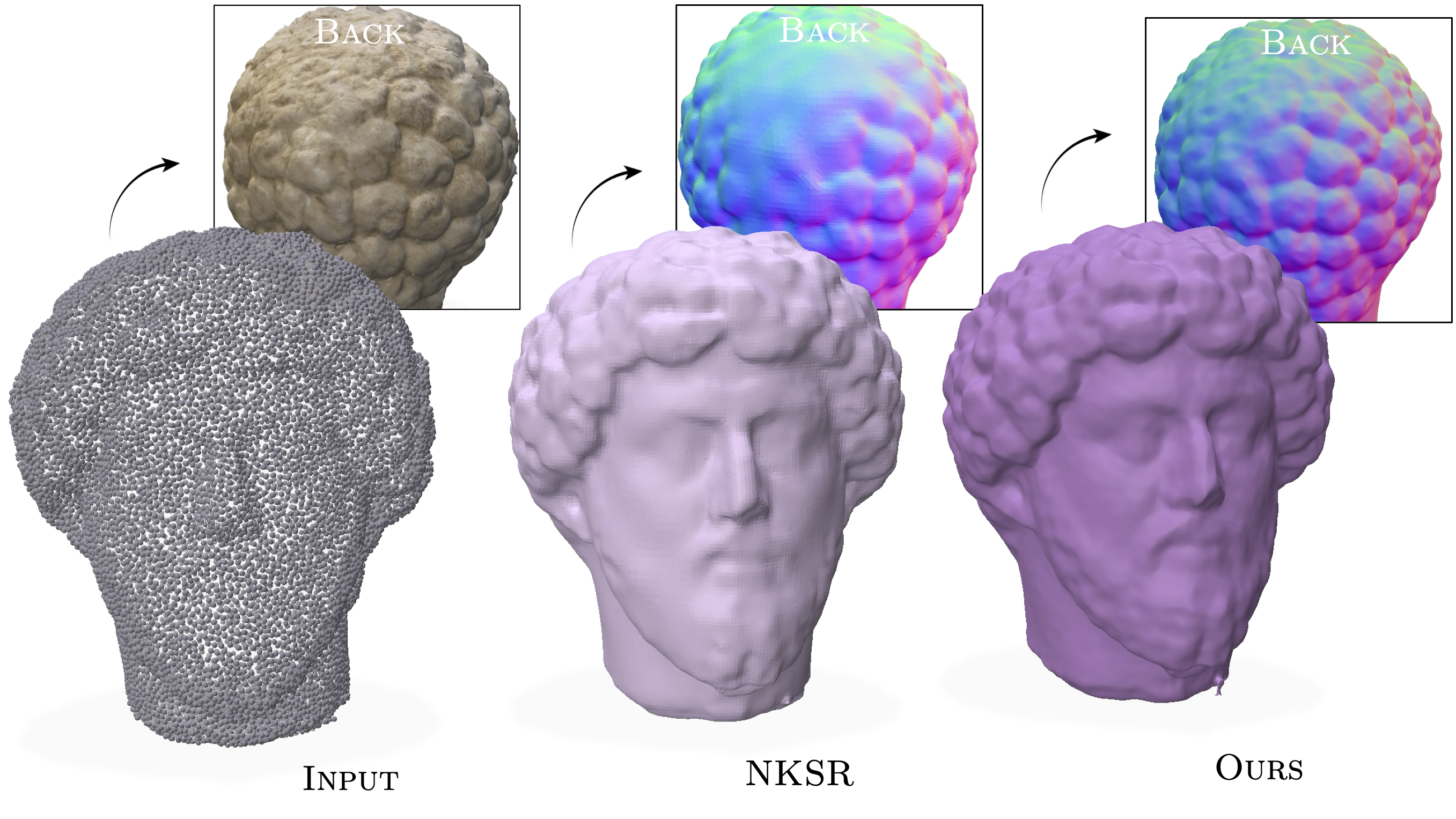}
    \caption{Comparison with NKSR \cite{huang2022neural} on the `Bust of Marcus Aurelius' \cite{fitzwilliam_head_2025}. Though NKSR is trained on a large 3D dataset and configured for maximum detail preservation, our method better captures fine surface details, particularly at the back of the head.}
    \label{fig:MarcusAureliusBust}
\end{figure}

A common strategy is to impose a global smoothness prior, as done in Poisson Surface Reconstruction (PSR) \cite{kazhdan2006poisson, kazhdan2013screened}, which solves a partial differential equation to produce globally smooth surfaces. Although effective for many objects and robust thanks to its global solution, approaches that rely solely on smoothness are limited: such priors often struggle to preserve sharp features and to leverage the intricate structural patterns and repetitive details common in real-world geometry. 
This motivates the development of priors capable of capturing richer geometric structure. The success of non-local methods in image processing \cite{buades2005non, dabov2007image}, which exploit self-similarity by connecting distant yet structurally alike patches, highlights the potential of extending this idea to geometry. Since real-world objects often exhibit strong internal repetition, self-similarity offers a compelling and intuitive geometric prior.\\

\begin{figure}[b]
    \centering
    \includegraphics[width=0.9\linewidth]{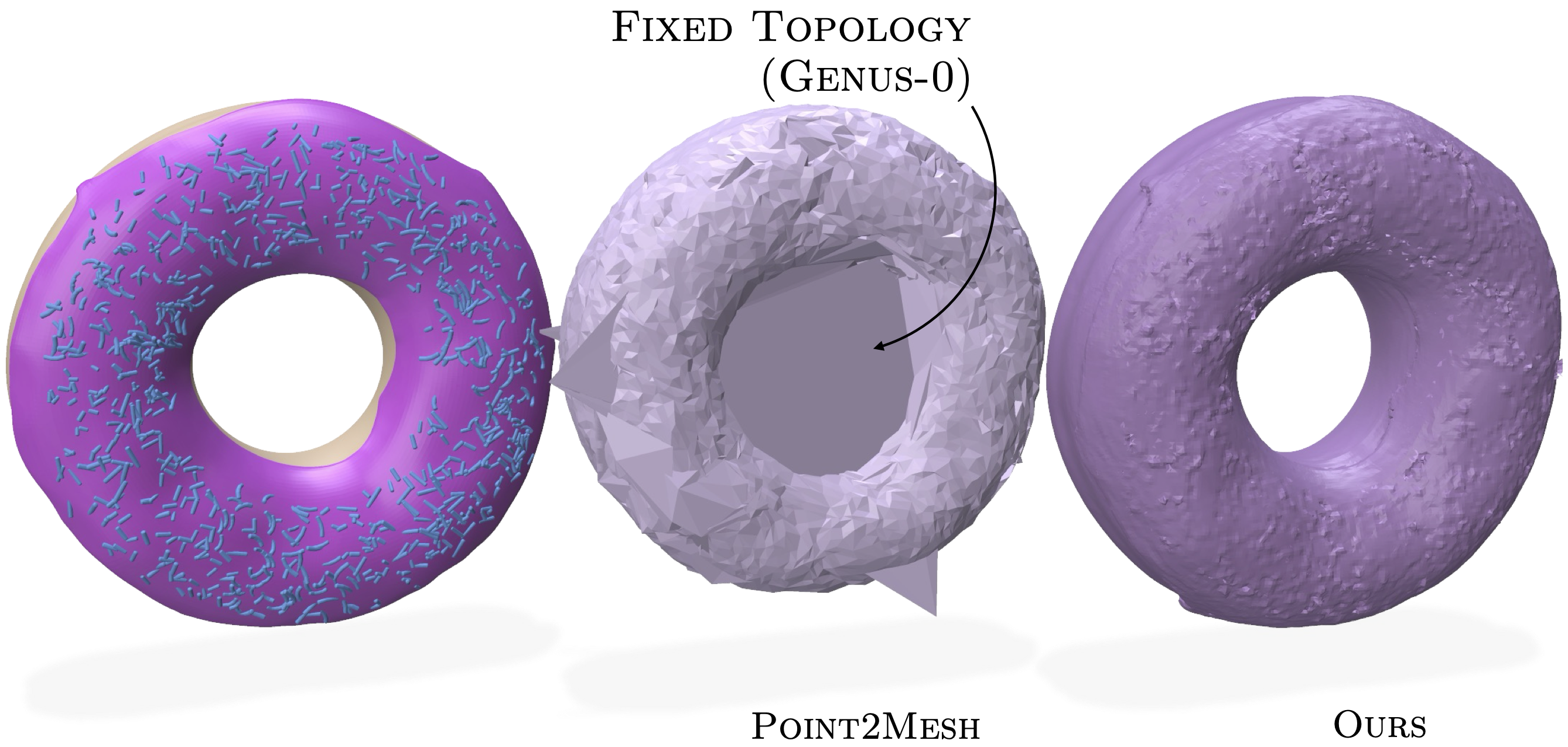}
    \caption{Deformation-based reconstruction methods, such as Point2Mesh, are limited by their reliance on a fixed input topology. In contrast, our \textit{implicit} approach provides greater flexibility in representing complex geometries.}
    \label{fig:topology}
\end{figure}

This naturally raises the question of how to effectively leverage self-similarity for 3D reconstruction. Point2Mesh \cite{hanocka2020point2mesh} addressed this by learning a self-prior directly on an explicit mesh, sharing MeshCNN kernel weights across the surface to predict displacement vectors that refine the mesh to fit the point cloud.
While this approach powerfully demonstrated the utility of learned self-similarity for capturing intricate details on a given mesh topology, its reliance on deforming an \textit{explicit} surface representation fundamentally limits topological flexibility (see Figure \ref{fig:topology}), can blur sharp features, and is less suited for arbitrary shapes or reconstructing directly from unoriented point clouds. This motivates the use of alternative representations, such as implicit functions, which are continuous and inherently support more flexible surface topologies.\\

While implicit surface representations, such as Signed Distance Functions (SDFs), offer a powerful and flexible framework for surface reconstruction, incorporating a learned self-prior into these models remains challenging. Classical approaches like Implicit Moving Least Squares (IMLS) \cite{shen2004interpolating, oztireli2009feature} excel at producing surface reconstructions but cannot exploit structural cues beyond point neighborhoods. More recently, neural implicit methods have emerged as a promising alternative, representing surfaces as continuous fields parameterized by neural networks. Several works have adapted these models for unstructured point cloud reconstruction without ground-truth SDFs, leveraging geometric constraints such as the Eikonal loss \cite{gropp2020implicit}, sign-agnostic supervision \cite{atzmon2020sal}, or gradient-based point projection \cite{ma2021neuralpull}. Others introduce local surface priors by operating on overlapping patches \cite{erler2020points2surf} or regularize the field using MLS-inspired smoothness terms \cite{wang2021neural}. While these approaches achieve impressive reconstructions and accommodate complex topologies, the priors they impose, whether based on smoothness, local patches, or global latent codes, are inherently limited in their ability to capture compact shape specific features (see Fig.~\ref{fig:MarcusAureliusBust}).\\ 

To bridge this gap, we draw inspiration from signal processing, where \textit{dictionaries} \cite{elad2006image} represent complex signals as sparse combinations of shared \textit{atoms}, effectively capturing structure by exploiting internal redundancy. This concept aligns powerfully with geometric self-similarity. We propose that a learnable dictionary can effectively encode such self-priors within a neural field. Attention mechanisms \cite{vaswani2017attention} provide a natural and flexible means to dynamically query and aggregate information from these dictionary elements to inform local surface predictions. Indeed, recent successes in generative 3D modeling, such as 3DShape2VecSet \citep{zhang20233dshape2vecset}, have demonstrated the power of set-latent approaches using attention for high-fidelity 3D shape representation, suggesting their promise for reconstruction tasks as well.\\



Building on learned self-priors and attention mechanisms for implicit representations, we introduce a novel approach for surface reconstruction. Our method learns a neural field that implicitly captures a distance field corresponding to the input point cloud, trained using self-supervised loss functions. Crucially, we leverage a learned dictionary and cross-attention to enable the field to recognize and exploit non-local structure, effectively achieving a form of spatial weight sharing. This learned self-prior allows the resulting distance field to guide the densification of sparse input regions. We show that this field can be further refined through a robust moving-least-squares (MLS) surface approximation, enabling the extraction of high-fidelity surfaces with rich geometric detail and flexible topology.
Our main contributions are: (1) a self-supervised implicit framework that learns a shape-specific geometric prior from the input point cloud via cross-attention to a learned dictionary, enabling non-local structure modeling; (2) demonstration that the learned field can be effectively refined using Robust Implicit MLS, yielding accurate, flexible surface reconstructions; and (3) state-of-the-art performance on self-similar shapes, highlighting the benefits of our attention-based self-prior.

\section{Related Work}
\label{sec:related_work}
\begin{figure*}
    \centering
    \includegraphics[width=\linewidth]{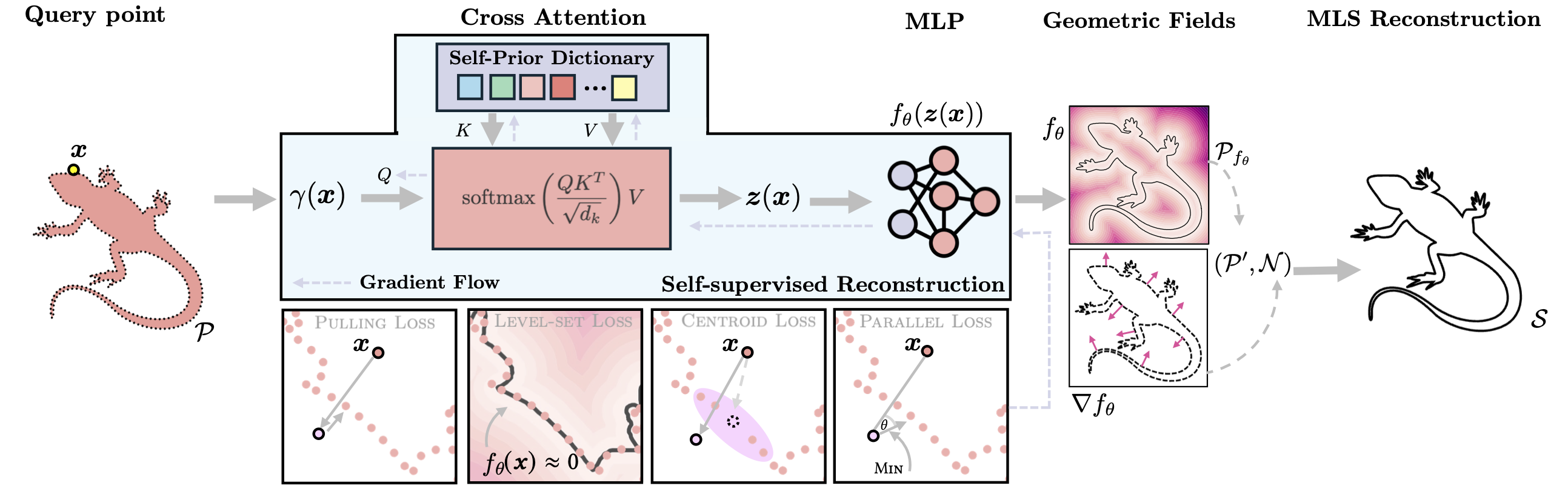} 
    \caption{We present a self-supervised surface reconstruction method based on a neural field conditioned via cross-attention. We assume access to an unoriented point cloud $\mathcal{P}$. Each input query point \( x \) is encoded via positional encoding \( \gamma(x) \) and interacts with a shared, learnable embedding dictionary to produce a latent representation \( z(x) \) that captures shape-specific geometric priors. An MLP then predicts geometric field, which we train with a number of geometric losses to recover the surface from the point cloud. Finally, Moving Least Squares (MLS) reconstruction is applied to refine the implicit surface.}
    \label{fig:main_overview_figure}
\end{figure*}
\paragraph{Classical Implicit Surface Reconstruction}
Reconstructing continuous surfaces from discrete point clouds is a foundational challenge in 3D vision and computer graphics \cite{berger2013benchmark}. Classical implicit methods typically represent surfaces as level sets of scalar functions, with Poisson Surface Reconstruction (PSR) being one of the most prominent examples \cite{kazhdan2006poisson, kazhdan2013screened}. PSR is well-regarded for producing smooth, watertight meshes when provided with high-quality normal estimates; however, its performance deteriorates in the presence of noisy data and may excessively smooth out fine details. Another influential approach is Implicit Moving Least Squares (IMLS) \cite{levin1998approximation, kolluri2008provably}, which defines an implicit function by locally fitting polynomials to the point cloud. IMLS is capable of preserving geometric detail and exhibits robustness to non-uniform sampling, but it remains heavily dependent on the availability of accurate and consistently oriented normals, a significant constraint when working with raw point clouds.

\paragraph{Neural Implicit Representations}
In recent years, deep neural networks have revolutionized 3D surface reconstruction. Neural implicit functions, such as signed distance fields (SDFs) and occupancy fields parameterized by MLPs, were first introduced as data-driven shape representations. Notable examples include DeepSDF \cite{park2019deepsdf} and Occupancy Networks \cite{mescheder2019occupancy}, which learn continuous volumetric functions by training on large shape datasets. While demonstrating the power of learned priors for complex topologies and shape interpolation, these early models typically rely on signed distances or occupancy labels for supervision, making them less robust to raw, unoriented, or noisy point clouds acquired in practice. Moreover, the learned shape spaces can struggle with out-of-distribution geometry, and the MLP representation itself can impose an implicit smoothness bias \cite{tancik2020fourier}.\\

To eliminate the need for ground-truth signed distances, a subsequent wave of methods trains neural implicit fields directly on the input point cloud in a self-supervised manner. Implicit Geometric Regularization (IGR) fits an MLP by encouraging it to vanish on input points and have unit gradient norm (Eikonal loss) \cite{gropp2020implicit}. Sign Agnostic Learning (SAL) devises loss functions invariant to the sign of the distance, enabling distance learning from raw unoriented points \cite{atzmon2020sal}. NeuralPull \cite{ma2020neural} introduced an explicit point-to-surface pulling loss, significantly improving reconstruction accuracy. While these self-supervised approaches (e.g., \cite{gropp2020implicit, atzmon2020sal, ma2020neural}) avoid large training sets, they mainly exploit local surface priors or collapse the shape into a single global latent code. As highlighted by \cite{hanocka2020point2mesh}, such priors may oversimplify or miss fine recurring details, as they do not explicitly model non-local self-similarity where repeating geometry could be leveraged.

\paragraph{Learned Geometric Priors for Reconstruction}
Recent advances have therefore explored more expressive learned priors. Some methods leverage external datasets: for instance, this can be done by pre-training models and then specializing them to new instances by optimizing query points \cite{ma2022surface}, or by specializing a decoder on local patches and projecting queries onto a learned surface via the decoder \cite{ma2022reconstructing}. Neural Kernel surface reconstruction (NKSR) \cite{huang2022neural} learns a multi-scale prior for fast reconstruction of large scenes. In contrast, we aim to learn a \textit{self-prior} derived from the input itself. Deep Geometric Prior (DGP) \cite{williams2021deep} fits an ensemble of small neural implicits to local patches, relying on the network's bias and consistency constraints, but lacks explicit global weight sharing. Point2Mesh \cite{hanocka2020point2mesh} explicitly learns a self-prior by optimizing a MeshCNN \cite{hanocka2019meshcnn} to deform an initial explicit mesh, effectively using shared convolutional kernels for self-similarity. Chu et al. \cite{chu2021unsupervised} adopt the deep prior paradigm, completing shapes from a single instance using a CNN interpreted through the neural tangent kernel (NTK) framework. Their method captures global self-similarity implicitly via shared features and NTK-informed architectural choices. However, it does not operate within a neural implicit surface framework or explicitly model reusable geometric patterns, leaving the explicit handling of global self-similarity in implicit representations largely underexplored.


\paragraph{Attention-Driven Shape Representations}
Our approach employs a learned dictionary of geometric tokens, accessed via an attention mechanisms, to implement a non-local self-prior. Attention has proven effective in enhancing the generalization of neural fields by conditioning them on sets of latent tokens \cite{vaswani2017attention, jiang2021cotr, sajjadi2022scene}. Notably, set-latent representations leveraging attention, such as 3DShape2VecSet \cite{zhang20233dshape2vecset}, have achieved high-fidelity 3D shape modeling by learning over collections of latent features.
Inspired by these advancements, we introduce a framework where a neural field, conditioned by a learnable dictionary of geometric tokens through cross-attention, implicitly learn a non-local self-prior. 

\section{Our Approach}
\label{sec:method}

We address the problem of reconstructing high-fidelity 3D surfaces from point clouds through a two-stage pipeline (see in Figure \ref{fig:main_overview_figure}). The core motivation behind our method is that self-similar patterns in local surface patches frequently recur across different regions of an object’s surface. Our method is designed to learn inherent self-prior and is able to infer missing or inaccurate geometric details.\\

\noindent
\textbf{(Stage 1) : Learning a neural distance field with implicit self-prior:} 
{We train an MLP $f_\theta$ to approximate the neural SDF field of the target shape. The self-prior is encoded within a compact learnable dictionary. For each query point $\boldsymbol{x}$, we perform cross-attention between its encoded position and dictionary entries to produce a feature representation $\boldsymbol{z}(\boldsymbol{x})$. We input $\boldsymbol{z}(\boldsymbol{x})$ to $f_\theta$ to predict its SDF.  The surface is defined as the zero-level set of the learned SDF field, while per-point normals follow from its spatial gradient.}\\

\noindent
\textbf{(Stage 2) : Geometric projection.}  
We discretize the learned geometric field and employ Robust Implicit Moving Least Squares (RIMLS) to define the final shape. This refinement step leverages the expressive capacity of $f_\theta$ alongside the feature-preserving properties of RIMLS, resulting in reconstructions that are both globally consistent and rich in detail.

\subsection{Dictionary-Conditioned Neural Field}
\label{ssec:self_prior_field}

A neural field is a continuous function, typically parameterized by a Multi-Layer Perceptron (MLP), that maps input coordinates to some target property. In our case, we aim to represent a continuous 3D shape by the level sets of an implicit distance function. Neural fields typically condition their predictions solely on spatial coordinates, requiring the network to reconstruct the entire shape from strictly local information. This localized perspective makes it challenging to capture long-range symmetries, repeated structures, and other global regularities that are implicitly shared across shapes. To overcome this limitation, we augment each point query via cross-attention to a shape-specific dictionary of learned embeddings. Because the same dictionary is accessible to all coordinates, the model can exchange information across distant regions and exploit the object’s structure.\\

\noindent
\textbf{Cross-Attention Dictionary}: ~More specifically, we adopt a decoder-only architecture, with an overview shown in Fig.~\ref{fig:main_overview_figure}. We begin by orthogonally initializing the embedding dictionary $\mathbf{E} \in \mathbb{R}^{N_k \times d_e}$, via QR decomposition of a random matrix~\cite{saxe2013exact}, to foster initial feature diversity and enhance learning stability. This dictionary consists of $N_k$ latent feature vectors, which are optimized jointly with the rest of the model parameters during training.\\

For a query point $\boldsymbol{x} \in \mathbb{R}^3$, we apply sinusoidal positional \cite{tancik2020fourier} encoding $\gamma(\boldsymbol{x})$ yielding a query vector $\gamma(\boldsymbol{x}) \in \mathbb{R}^{d_q}$. We then linearly project the query position to yield $\textbf{q} = W_{\gamma}\gamma(\boldsymbol{x})$, such that $\textbf{q} \in \mathbb{R}^{d_e}$. To perform cross-attention between the query position $\boldsymbol{x}$ and the embedding dictionary $\textbf{E}$, we apply \textit{multi-headed attention} (MHA) with $H$ heads. For each head $h = 1, \dots, H$, the query, keys, and values are linearly projected via learned matrices $(\mathbf{W}_q^{(h)}, \mathbf{W}_k^{(h)}, \mathbf{W}_v^{(h)})$ such that:
\begin{equation}
\mathbf{Q}_h = \mathbf{q}\mathbf{W}_q^{(h)}, \quad \mathbf{K}_h = \mathbf{E}\mathbf{W}_k^{(h)}, \quad \mathbf{V}_h = \mathbf{E}\mathbf{W}_v^{(h)}.
\end{equation}
Scaled dot-product attention is computed independently for each head:
\begin{equation}
\text{Attn}(\mathbf{Q}_h, \mathbf{K}_h, \mathbf{V}_h) = \text{softmax}\left(\frac{\mathbf{Q}_h \mathbf{K}_h^\top}{\sqrt{d_k}}\right)\mathbf{V}_h,
\end{equation}
where $d_k$ is the key dimensionality per head. The outputs from all heads are concatenated and projected through $\mathbf{W}_o$ to produce the final context vector $\boldsymbol{z}(\boldsymbol{x}) \in \mathbb{R}^{d_{\text{out}}}$:
\begin{equation}
\boldsymbol{z}(\boldsymbol{x}) = \text{Concat}(\text{Attn}_1, \dots, \text{Attn}_H) \mathbf{W}_o.
\end{equation}

\noindent
\textbf{Signed-distance Prediction Head}: To preserve fine-grained spatial information, we introduce a learned linear projection of the raw coordinates, defined as \( \tilde{\boldsymbol{x}} = W_{\mathrm{proj}} \boldsymbol{x} \), where \( \tilde{\boldsymbol{x}} \in \mathbb{R}^{d_{\mathrm{out}}} \). This projected signal is added to the context vector, yielding the final input \( \bar{\boldsymbol{z}}(\boldsymbol{x}) = \boldsymbol{z}(\boldsymbol{x}) + \tilde{\boldsymbol{x}} \).
We consider a multilayer perceptron (MLP) \( f_\theta : \mathbb{R}^{d_{\mathrm{out}}} \to \mathbb{R} \) with \( L \) hidden layers of width \( d_{\mathrm{hid}} \) and ReLU activations. To ease optimization in deeper networks, we follow \cite{li2023neuralgf} and add a single skip connection by concatenating the original input to the intermediate representation after the \( \lfloor L/2 \rfloor \)-th layer. 
Following \cite{atzmon2020sald,gropp2020implicit}, we adopt geometric initialization to encourage signed distance function (SDF) behavior. Specifically, (i) weights in the hidden layers are initialized from a normal distribution \( \mathcal{N}(0,\, 2/d_{\mathrm{hid}}) \); and (ii) the final layer is initialized with zero bias and small weights, ensuring outputs are near zero at initialization and promoting stable training.

\subsection{Training the Neural Field}
\label{sec:training}

We denote the full attentive neural field $f_\theta(\bar{\boldsymbol{z}}(\boldsymbol{x}))$ by $g_{\phi}(\boldsymbol{x})$.  
Following \cite{ma2020neural, li2023neuralgf}, training uses two complementary supervision sets derived from the input point cloud $\mathcal{P}=\{\mathbf{p}_i\}_{i=1}^N$:  
off-surface points $\mathcal{Q}$, obtained by adding Gaussian noise $\boldsymbol{\delta} \sim \mathcal{D}$ to uniformly sampled $\mathbf{p} \in \mathcal{P}$,  
and on-surface points $\mathcal{G}$, directly subsampled from $\mathcal{P}$ to anchor the zero-level set.  
The complete training set is $\mathcal{T} = \mathcal{Q} \cup \mathcal{G}$. We estimate the unit normal:  
\[
\boldsymbol{\nu}(\boldsymbol{x}) = \frac{\nabla_{\boldsymbol{x}}g_{\phi}(\boldsymbol{x})}{\|\nabla_{\boldsymbol{x}} g_{\phi}(\boldsymbol{x})\|},
\]
and define the projection operator:
\[
\mathscr{P}(\boldsymbol{x}) = \boldsymbol{x} - g_{\phi}(\boldsymbol{x}) \cdot \boldsymbol{\nu}(\boldsymbol{x}),
\]
which moves points toward the surface; $\mathscr{P}_m$ denotes $m$ successive applications. We employ established geometric losses from prior neural distance function works \cite{ma2020neural, zhou2023learning, li2023neuralgf}:\\

\noindent
\textbf{Global Surface Loss}: Enforces that projected points lie close to surface samples:
\[
\mathcal{L}_\alpha = \mathbb{E}_{\mathbf{q} \sim \mathcal{Q}}\!\left[\|\mathscr{P}_2(\mathbf{q}) - \hat{\mathbf{q}}\|^2\right] +
\mathbb{E}_{\mathbf{g} \sim \mathcal{G}}\!\left[\|\mathscr{P}_2(\mathbf{g}) - \mathbf{g}\|^2\right],
\]
where $\hat{\mathbf{q}}$ is the nearest neighbor of $\mathbf{q}$ in $\mathcal{P}$.  
This pulls off-surface points toward the data and anchors on-surface points.\\

\noindent
\textbf{Level-set Loss}: Encourages $g_\phi \approx 0$ for on surface points and also after one applications of the projection operator:
\[
\mathcal{L}_{\beta} = \mathbb{E}_{\mathbf{g} \sim \mathcal{G}}\!\left[g_\phi(\mathbf{g})^2\right] +
\mathbb{E}_{\mathbf{t} \sim \mathcal{T}}\!\left[g_\phi(\mathscr{P}(\mathbf{t}))^2\right].
\]
The first term constrains known surface samples and the second regularises refined points.\\

\noindent
\textbf{Local Displacement Loss}: Aligns predicted displacements with local geometric estimates across scales $s$:
\[
\mathcal{V}_s(\mathbf{q}) = \mathbf{q} - \frac{1}{K_s} \sum_{k=1}^{K_s} \mathbf{p}_k.\]
Here, $\mathcal{V}_s(\mathbf{q})$ denotes the displacement from $\mathbf{q}$ to the centroid of its $K_s$ nearest neighbours in $\mathcal{P}$. The local displacement loss is:
\[
\mathcal{L}_\gamma = \sum_s \mathbb{E}_{\mathbf{q} \sim \mathcal{Q}}
\left[\|(\mathbf{q} - \mathscr{P}_1(\mathbf{q})) - \mathcal{V}_s(\mathbf{q})\|^2\right],
\]
which promotes consistency across varying point densities.\\

\noindent
\textbf{Normal Consistency Loss}: Encourages stable surface normals during refinement:
\[
\mathcal{L}_\delta = \mathbb{E}_{\mathbf{x}\sim\mathcal{T}}\!\left[ w(\mathbf{x}) \cdot
\big(1 - d_{\cos}(\boldsymbol{\nu}(\mathbf{x}), \boldsymbol{\nu}(\mathscr{P}(\mathbf{x})))\big)\right],
\]
with $w(\mathbf{x})=\exp(-\rho |g_\phi(\mathbf{x})|)$.  
The loss enforces minimal change in normals as points move closer to the surface.

\paragraph{Total Loss}  
The training objective is a weighted sum of all terms:
\[
\mathcal{L} = \alpha\,\mathcal{L}_\alpha + \beta\,\mathcal{L}_\beta + \gamma\,\mathcal{L}_\gamma + \delta\,\mathcal{L}_\delta,
\]
where $\alpha,\beta,\gamma,\delta$ control the relative influence of global alignment, level-set consistency, local displacement, and normal smoothness, respectively.

\begin{figure}[!b]
    \centering
    \includegraphics[width=\linewidth]{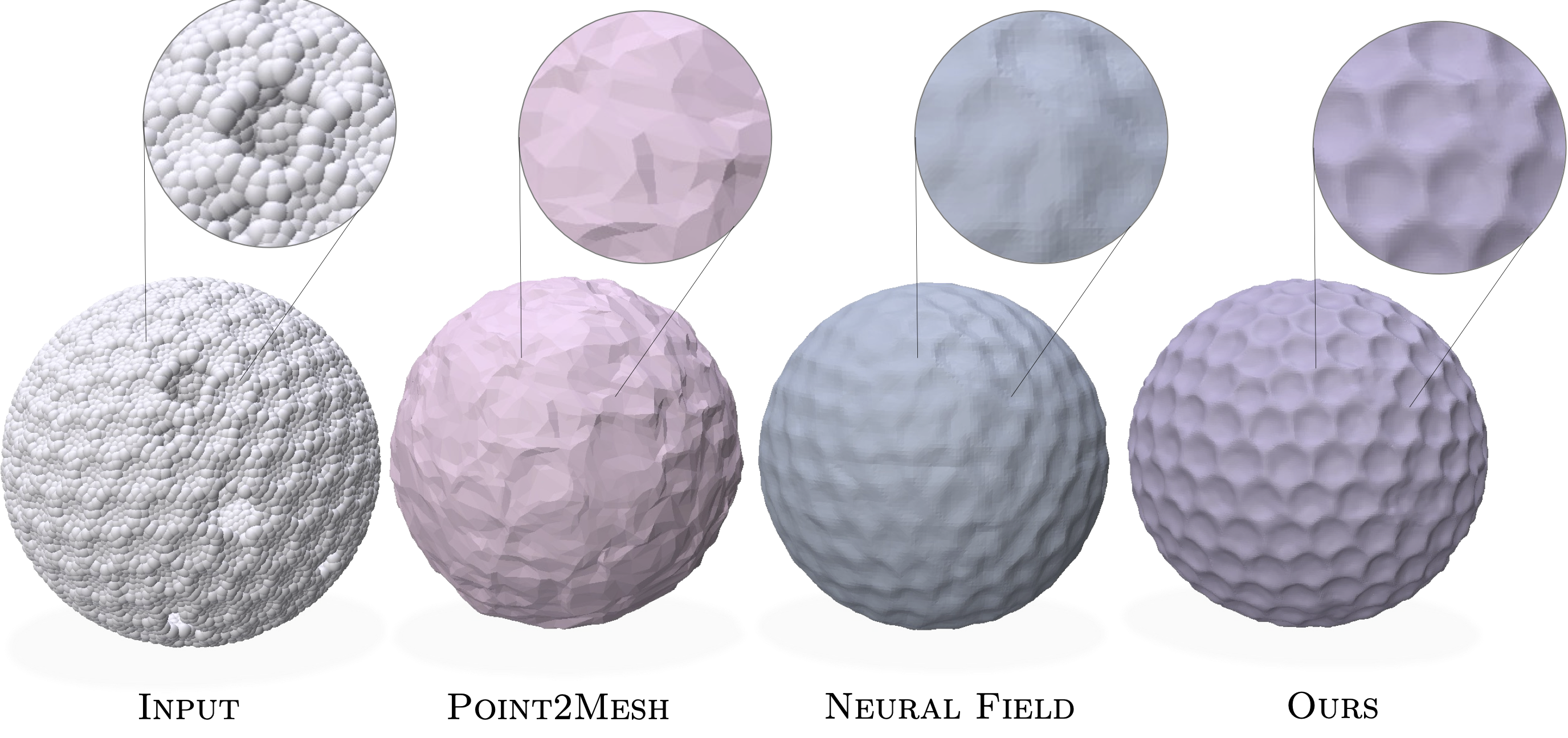}
    \caption{Our method combines a learned self-prior with explicit point cloud control to preserve surface detail, outperforming approaches that rely solely on learned priors (Point2Mesh) or neural fields without attention.}
    \label{fig:selfprior_golfball}
\end{figure}
\subsection{Geometric Quantity Estimation}
We use the learnt implicit field $g_\phi$ both to lightly inpaint sparse regions and to obtain normals; this allows the self-prior to guide the surface reconsturction as in Fig.~\ref{fig:selfprior_golfball}. 

\paragraph{Inpainting}
We extract the zero level set of $g_\phi$ with Marching Cubes~\cite{lorensen1998marching} and uniformly sample it to obtain a dense auxiliary set $\tilde{\mathcal{P}}$. For each $\tilde{\boldsymbol{p}}_j\in\tilde{\mathcal{P}}$, let
$d(\tilde{\boldsymbol{p}}_j,\mathcal{P})=\min_{\boldsymbol{p}\in\mathcal{P}}\|\tilde{\boldsymbol{p}}_j-\boldsymbol{p}\|_2$
be its nearest-neighbour distance to the input cloud $\mathcal{P}$ (i.e., a one-sided Chamfer term). With $\sigma_d$ the standard deviation of $\{d(\tilde{\boldsymbol{p}}_j,\mathcal{P})\}$, we keep only far points
$\mathcal{P}_{\text{fill}}=\{\tilde{\boldsymbol{p}}_j~|~d(\tilde{\boldsymbol{p}}_j,\mathcal{P})\ge 3\sigma_d\}$.
The augmented set is then given by $\mathcal{P}'=\mathcal{P}\cup\mathcal{P}_{\text{fill}}$.

\paragraph{Normals}
For each $\boldsymbol{p}_i\in\mathcal{P}'$, we estimate a normal by the normalized gradient of the field,
$\boldsymbol{n}_i=\nabla g_\phi(\boldsymbol{p}_i)/\|\nabla g_\phi(\boldsymbol{p}_i)\|_2$, and denote the full set of normals as $\mathcal{N}=\{\boldsymbol{n}_i\}$.

\paragraph{MLS Refinement}
To move beyond the stability–detail trade-off inherent in direct reconstruction from sparse, uneven point clouds, we refine the surface using Robust Implicit Moving Least Squares (RIMLS)~\cite{oztireli2009feature}. The field $g_\phi$ first inpaints gaps and provides coherent normals, yielding $(\mathcal{P}',\mathcal{N})$ as an enhanced point set. RIMLS then reconstructs the surface while preserving sharp features and fine detail, guided by these refinements. Finally, we evaluate the implicit function on a grid and extract the mesh using Marching Cubes~\cite{lorensen1998marching}.

\section{Experiments}
\label{sec:experiments}

\newcolumntype{Y}{>{\centering\arraybackslash}X}
\newcommand{\best}[1]{\textbf{\textcolor{green!60!black}{#1}}}
\newcommand{\bestcell}[1]{\cellcolor{green!15}#1}

\newcommand{\secondbestcell}[1]{\cellcolor{yellow!15}#1}

We evaluate of our method through a series of qualitative and quantitative experiments involving shapes with low density regions, noise, and different topologies. We provide implementation details in section \ref{experimental_details} of  the supplementary materials.\\

\subsection{Experimental Setup}\label{sec:reconstruction}

\noindent
\textbf{Datasets}: We evaluate our method on four datasets. First, we use the Surface Reconstruction Benchmark (SRB) \cite{berger2013benchmark}, which contains five range-scan models and is a standard dataset for surface reconstruction. Second, we curate a set of objects with strong self-similarity to assess performance on inputs with repeated structure. Third, we test robustness using a subset of Thingi10K \cite{zhou2016thingi10k}, specifically a variant containing noise from \cite{erler2020points2surf}, which include Gaussian noise to simulate sensor imperfections. Finally, we evaluate on the full set of models provided in the public release of Point2Mesh \cite{hanocka2020point2mesh}.\\



\noindent

\noindent
\textbf{Comparison}: We evaluate our method against a broad spectrum of surface reconstruction techniques, including analytical, optimization-based, and learning-driven approaches. Analytical baselines include Screened Poisson Surface Reconstruction (SPSR) \cite{kazhdan2013screened}, which produces smooth, complete surfaces under clean input conditions, but is sensitive to noise and requires oriented normals, and Diffusing Winding Gradients (DWG) \cite{liu2025diffusing}, a recent non-learning method that reconstructs from unoriented point clouds via diffusion of generalized winding number gradients, offering strong scalability but lacking learned priors. Optimization-based methods such as Shape-as-Points (SAP) \cite{peng2021shape} formulate classical objectives as differentiable losses, minimizing Chamfer distance in a Poisson-inspired setting. Learning-based approaches include Neural Kernel Surface Reconstruction (NKSR) \cite{huang2022neural}, which learns transferable shape priors; Point2Mesh (P2M) \cite{hanocka2020point2mesh}, which learns per-shape self-priors through mesh deformation with fixed connectivity; and other techniques like Deep Geometric Prior (DGP) \cite{williams2021deep}, Neural-IMLS (NIMLS) \cite{wang2021neural}, and Predictive Context Prior (PCP) \cite{ma2022surface}, which differ in supervision and prior modeling strategies. We also compare against recent neural field–based methods, including PG-SDF \cite{koneputugodage2024small} and Neural Singular Hessian \cite{wang2023neural}, which define implicit surfaces using continuous neural fields trained directly on point clouds.\\


\subsection{Experimental Results}

\noindent
\textbf{SRB Dataset}: Table~\ref{table:srb_metrics} reports the reconstruction metrics of our method on the SRB dataset. Our approach achieves state-of-the-art performance across multiple evaluation criteria. Notably, it yields the lowest Chamfer Distance and Hausdorff Distance, indicating superior geometric accuracy and surface completeness. Qualitative reconstruction results are provided in Sec.~\ref{QualitativeSRB} of the appendix.\\

\begin{table}[!ht]
\centering
\small
\caption{Surface reconstruction metrics on the SRB dataset, proposed in \cite{berger2013benchmark}. Lower is better for CD and HD; higher is better for NC and F-score.}
%

\label{table:srb_metrics}
\renewcommand\theadfont{\bfseries}
\setlength{\tabcolsep}{3pt}
\begin{tabularx}{\linewidth}{@{}l|YYYY}
\toprule
\textsc{\textbf{Method}} 
&\textsc{\thead{CD~(↓)}} & \textsc{\thead{HD~(↓)}} & \textsc{\thead{NC~(↑)}} & \textsc{\thead{FS~(↑)}}\\
\midrule
SPSR  & 
0.413&	1.498&	0.919 &	71.63 \\
DGP & 
0.022 & 0.701 & \secondbestcell{0.951} & \secondbestcell{75.67} \\
P2M &
0.177 & 0.902 & 0.857 & 24.47   \\
PCP &
0.283 & 2.039 & 0.900 & 49.39 \\
NIMLS &
0.283 & 1.992 & 0.913 & 54.62 \\
NKSR  & 
\secondbestcell{0.019} &  \secondbestcell{0.614} & 0.949 & \bestcell{75.98} \\
SAP & 
0.024 &0.682 & 0.936	& 75.49\\
\hline
\textbf{Ours} & 
\bestcell{0.016} &	\bestcell{0.484} &	\bestcell{0.956}&	75.54\\
\bottomrule
\end{tabularx}
\end{table}

\begin{figure}[!h]
    \centering
    \includegraphics[width=\linewidth]{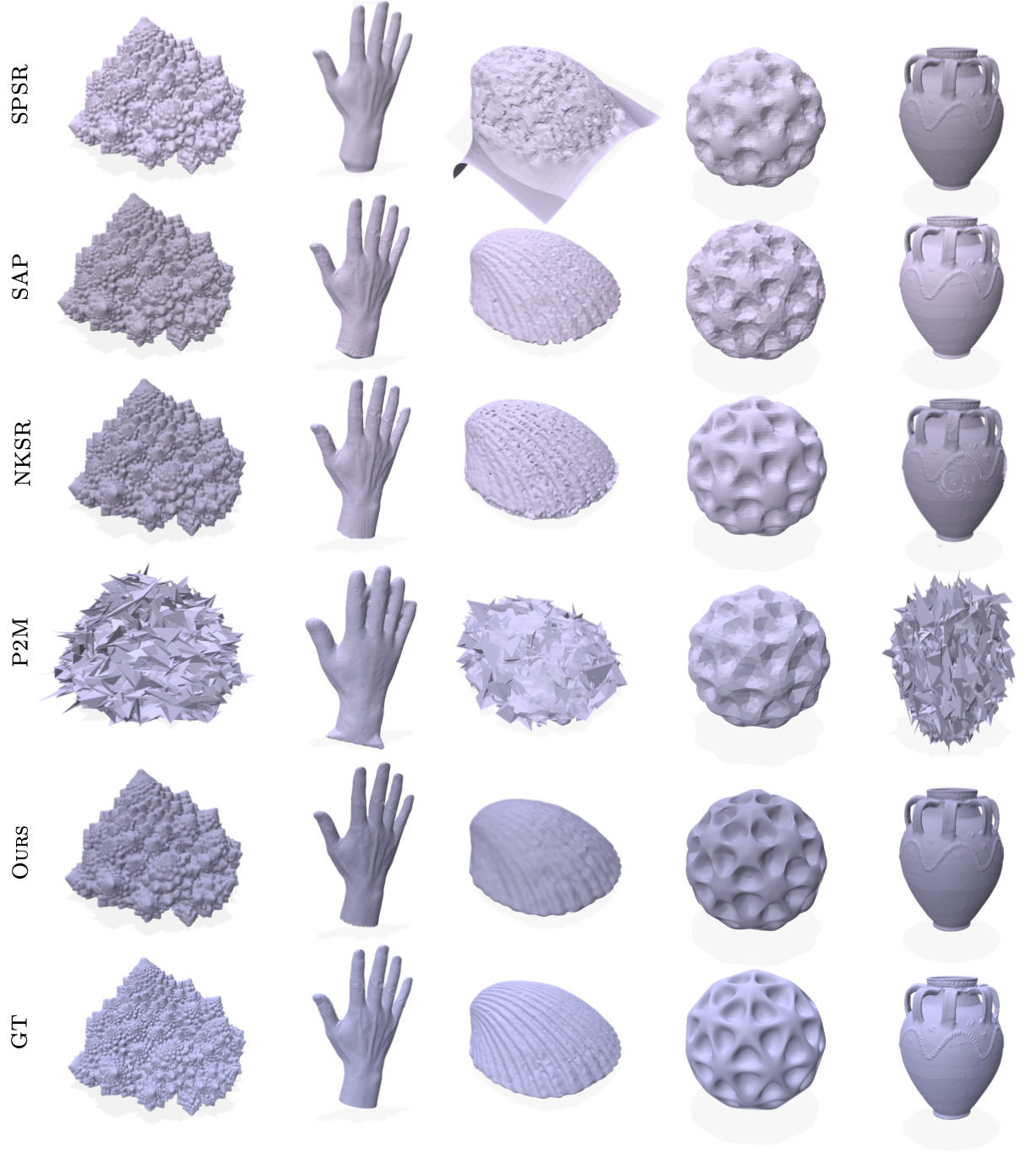}
    \caption{We present qualitative comparison between our method and other leading reconstruction methods on shapes with high amounts of self-similarity; Our approach excels at capturing global shape properties while retaining local shape details.}
    \label{fig:qual_compare}
\end{figure}

\noindent
\textbf{Self-similar Dataset}: Table~\ref{table:self-sim_metric} further demonstrates the effectiveness of our method in reconstructing shapes characterized by strong self-similarity. Our approach achieves the best performance across Chamfer Distance, Normal Consistency, and F-score, highlighting its robustness and accuracy in challenging reconstruction scenarios. Qualitative comparisons are shown in Fig.~\ref{fig:qual_compare}.
\begin{table}[!ht]
\centering
\small
\caption{Surface reconstruction metrics on our dataset consisting of objects with large self-similarity. Lower is better for CD and HD; higher is better for NC and F-score.}
\label{table:self-sim_metric}
\renewcommand\theadfont{\bfseries}
\setlength{\tabcolsep}{3pt}
\begin{tabularx}{\linewidth}{@{}l|YYYY}
\toprule
\textsc{\textbf{Method}} & \textsc{\thead{CD~(↓)}} & \textsc{\thead{HD~(↓)}} & \textsc{\thead{NC~(↑)}} & \textsc{\thead{FS~(↑)}}\\
\midrule
SPSR  & 0.248 & 1.475 & 0.866 & 61.89 \\
SAP   & 0.021 & 0.690 & \secondbestcell{0.906} & \secondbestcell{71.21} \\
NKSR  & \secondbestcell{0.019} & {0.512} & 0.897 & 68.56 \\
P2M   & 0.239 & 1.384 & 0.695 & 19.24 \\
NSH & 0.043 & 0.971 & 0.859 & 61.95 \\
WDG  & 0.038 & 1.246 & 0.785 & 48.39 \\
PG-SDF & \secondbestcell{0.019} & \bestcell{0.397} & 0.872 & 66.38 \\
\hline
\textbf{Ours} & \bestcell{0.017} & \secondbestcell{0.438} & \bestcell{0.907} & \bestcell{72.44} \\
\bottomrule
\end{tabularx}
\end{table}

\begin{figure}[!ht]
    \centering
    \includegraphics[width=\linewidth]{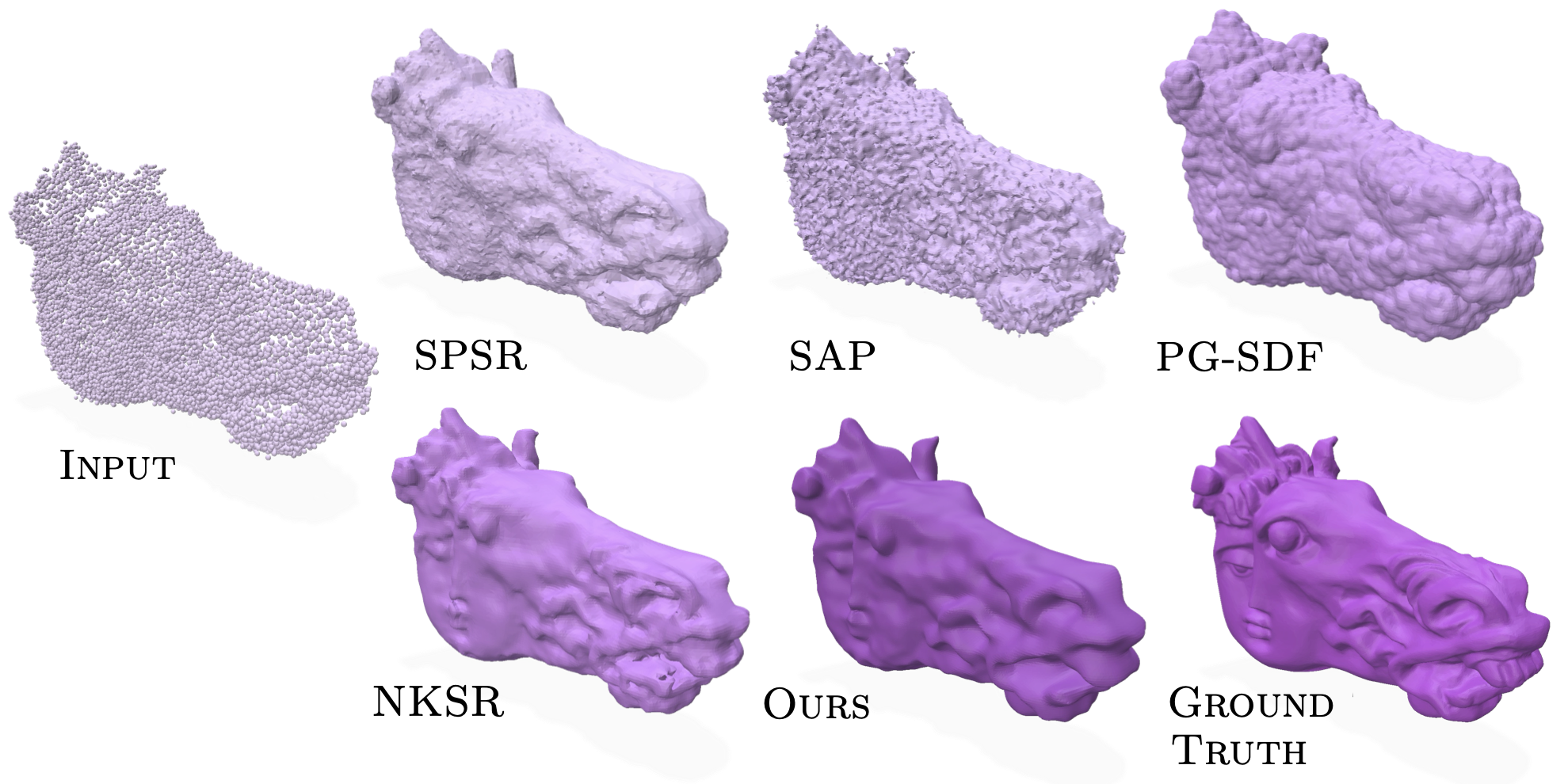}
    \caption{Comparison on a Thinki10K sample with added Gaussian noise. Even with noise, our method preserves fine-grained details and similar patterns better than competing approaches.}
    \label{fig:Noised}
\end{figure}

\noindent
\textbf{Thingi10K Dataset}: Table~\ref{table:thingi10k} highlights the robustness of our method on the noised Thingi10K dataset. While NKSR achieves the top performance due to explicit noise-aware training, our approach ranks second in Chamfer Distance and Normal Consistency, demonstrating strong resilience to noise and sparsity. As shown in Fig.~\ref{fig:Noised}, our method effectively suppresses input noise while preserving fine geometric details.\\

\begin{table}[h]
\centering
\small
\caption{Surface reconstruction metrics over samples from the Thingi10K dataset. Lower is better for CD and HD; higher is better for NC and F-score.}
\label{table:thingi10k}
\renewcommand\theadfont{\bfseries}
\setlength{\tabcolsep}{3pt}
\begin{tabularx}{\linewidth}{@{}l|YYYY}
\toprule
\textsc{\textbf{Method}} & \textsc{\thead{CD~(↓)}} & \textsc{\thead{HD~(↓)}} & \textsc{\thead{NC~(↑)}} & \textsc{\thead{FS~(↑)}}\\
\midrule
SPSR  & 0.032 & 0.620 & 0.899 & \secondbestcell{67.31}\\
SAP   & 0.022 & \secondbestcell{0.385} & 0.734& 62.77 \\
NKSR  & \bestcell{0.019} & \bestcell{0.398} & \bestcell{0.939} & \bestcell{70.50} \\
PG-SDF & 0.151 & 0.596 & 0.860 & 1.72  \\
\hline
\textbf{Ours} & \secondbestcell{0.021} & 0.458 & \secondbestcell{0.931} & 63.71 \\
\bottomrule
\end{tabularx}
\end{table}

\noindent
\textbf{Point2Mesh Dataset}: Since Point2Mesh is the most directly comparable method in terms of learning a self-prior for surface reconstruction, we further evaluate our approach on the publicly available dataset provided by its authors. The results, shown in Fig.~\ref{fig:p2m_dataset} of the appendix, indicate that while both methods capture the underlying geometry, our approach produces smoother surfaces and better preserves sharp features.\\

\begin{figure}
    \centering
    \includegraphics[width=0.95\linewidth]{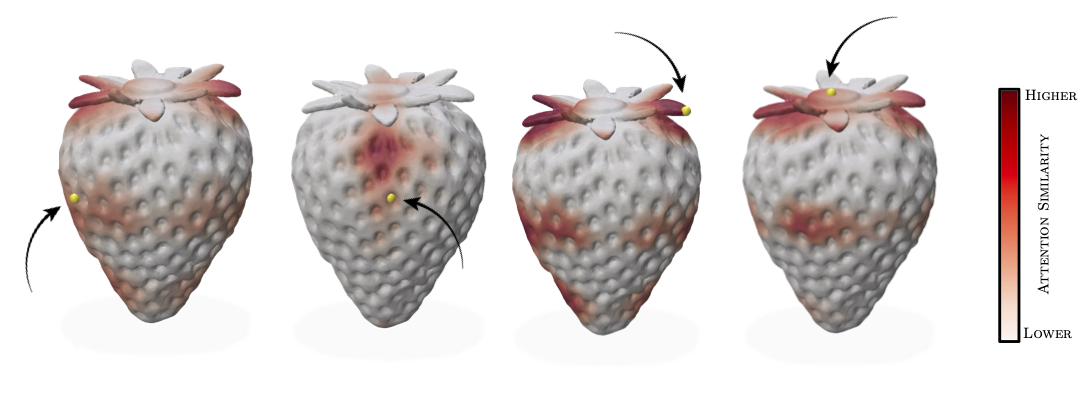}
    \caption{Attention weight similarity across the surface, relative to a yellow query point. Similarity (white: low, red: high) is computed via dot product of attention weights.}
    \label{fig:attentionstraw}
\end{figure}

\noindent
\textbf{Interpretability}:
To better understand the behavior of our attention mechanism, we visualize the attention weight similarity across the surface relative to a selected query point. We train our model on the strawberry point cloud with a dictionary size of 16. In Fig.~\ref{fig:attentionstraw}, the similarity is measured via the dot product of attention weights, where warmer colors (red) indicate higher similarity and cooler colors (white) denote lower similarity. We observe that the model learns to couple non-local regions; for example, when the query is located on or near the leaf structure, the attention shifts to highlight other leaf regions, despite their spatial separation. This behavior indicates that the model learns a meaningful self-prior, effectively linking similar but spatially distant regions of the shape.\\

\noindent\textbf{Ablation studies}:
\label{sec:ablation}
We evaluate the contributions of the RIMLS refinement and the attentive dictionary through ablation studies on the self-similar dataset. To this end, we compare three configurations: (i) the full model, (ii) a variant without RIMLS refinement, and (iii) a variant without both RIMLS and attention (No MLS + No Attn); for the latter we increase the parameter count of the neural field to approximately match the full model’s parameter count.  Quantitative results for all configurations are reported in Table~\ref{table:ablation_self_sim}. The base model without attention or MLS refinement has the weakest performance across all metrics. Introducing attention substantially improves reconstruction quality, while the addition of RIMLS refinement yields further gains, demonstrating the complementary benefits of both components.\\ 



\newcommand{\cmark}{\textcolor{black}{\ding{51}}}

\newcommand{\xmark}{\textcolor{red}{\ding{55}}}
\begin{table}[h]\label{ablation_study}
\centering
\small
\caption{Ablations on the self-similar dataset. Lower is better for CD and HD; higher is better for NC and F-score.}
\label{table:ablation_self_sim}
\renewcommand\theadfont{\bfseries}
\setlength{\tabcolsep}{3pt}
\begin{tabularx}{\linewidth}{@{}cc|c|YYYY}
\toprule
\textsc{Attn} & \textsc{MLS} & \textsc{Params}&
\textsc{\thead{CD~(↓)}} & \textsc{\thead{HD~(↓)}} &
\textsc{\thead{NC~(↑)}} & \textsc{\thead{FS~(↑)}} \\
\midrule
\xmark & \xmark &1.18 M &0.021 & 0.534 & 0.876 & 66.62 \\
 \cmark & \xmark &1.16 M  &0.019 & 0.494 & 0.903 & 67.52 \\
\cmark &\cmark &1.16 M  &\textbf{0.017} & \textbf{0.438} & \textbf{0.907} & \textbf{72.44} \\
\bottomrule
\end{tabularx}
\end{table}

\begin{figure}[!h]
    \centering
    \includegraphics[width=0.75\linewidth]{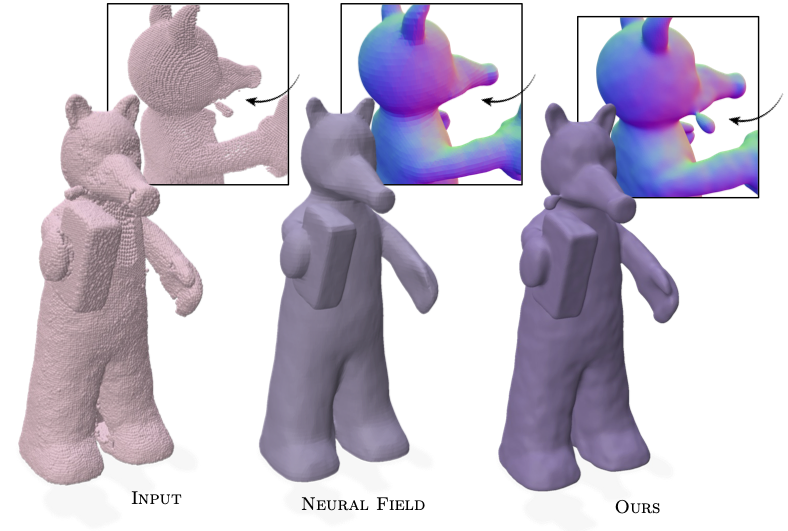}
    \caption{The zero level set of the attentive neural field may miss fine geometric details (e.g., the missing tube structure), but its gradient field still captures meaningful surface normals. Left: input point cloud. Middle: zero level set. Right: our hybrid method uses gradients for more complete reconstruction.}
    \label{fig:rimls_compare}
\end{figure}

Figure~\ref{fig:rimls_compare} illustrates the effect of the MLS refinement. In this example, the raw zero-level set of the neural field fails to capture a thin pipe structure. By applying the refinement, the structure is successfully recovered, demonstrating its ability to preserve fine geometric details.

\subsection{Normal Estimation Experiments}
While designed for surface reconstruction, we evaluate our method on surface normal estimation using the PCPNet dataset, following the protocol of \cite{li2023neuralgf}. The results highlight the robustness of our approach across different noise levels and point cloud densities. Detailed comparisons and additional results are provided in Sec.~\ref{appendix:nml_est} of the appendix.


\section{Conclusion}
\label{sec:conclusion}

We introduced a self-supervised approach for high-fidelity point cloud reconstruction, leveraging an implicit attention prior. The method learns a shape-specific prior directly from the input by training an implicit neural field conditioned on a learnable dictionary of geometric tokens via cross-attention. This enables the network to capture non-local self-similarities and repeating structural patterns without external training data, guiding both sparse-region densification and high-quality analytic normal estimation. These features are integrated into a robust implicit moving least squares (RIMLS) framework, combining the global structural awareness of the learned prior with the local accuracy of classical reconstruction. Experiments suggest that our self-prior demonstrates competitive performance, showing strengths in detail preservation, topological adaptability, and robustness to noise and sparsity compared to both classical and learning-based methods. By learning complex, shape-specific priors from input alone, our approach overcomes key limitations of traditional methods and provides a flexible foundation for challenging scenarios. Future directions include extensions to dynamic or large-scale scenes, transfer learning between shapes, and generative modeling of novel shapes that inherit the structural traits of a reference.

\newpage
\clearpage
{
    \small
    \bibliographystyle{ieeenat_fullname}
    \bibliography{reference}
}
\clearpage
\setcounter{page}{1}
\maketitlesupplementary

\section{Further Experiments}
\subsection{Normal Estimation}\label{appendix:nml_est}

\noindent
\textbf{Dataset and Metric}: We adopt the evaluation protocol from \cite{li2023neuralgf}, using the PCPNet dataset \cite{guerrero2018pcpnet}, which contains synthetic 3D shapes with a variety of surface characteristics, ranging from smooth regions to complex geometries with sharp features. Each shape is provided as a clean point cloud along with versions corrupted by Gaussian noise at three levels (0.12\%, 0.6\%, and 1.2\% of the bounding box diagonal), as well as point clouds with non-uniform densities. Following \cite{li2023neuralgf}, we report the oriented root mean square error (RMSE) of predicted normals (see Appendix~\ref{appendix:normal_estimation} for details). Baseline methods and their corresponding results are adopted from \cite{li2023neuralgf} to ensure consistency and comparability.\\

\noindent
\textbf{Comparison}
We evaluate against a comprehensive set of baselines, including both classical and learning-based methods. Classical techniques include Principal Component Analysis (PCA) \cite{hoppe1992surface} and Locally Robust Regression (LRR) \cite{zhang2013point}, each combined with three orientation propagation strategies: Minimum Spanning Tree (MST) \cite{hoppe1992surface}, Sign Orientation Propagation (SNO) \cite{schertler2017towards}, and Orientation Determination Propagation (ODP) \cite{metzer2021orienting}. Learning-based baselines include AdaFit \cite{zhu2021adafit}, HSurf-Net \cite{li2022hsurf}, PCPNet \cite{guerrero2018pcpnet}, SHS-Net \cite{li2023shs}, and NeuralGF \cite{li2023neuralgf}.\\

\noindent
\textbf{Results.}
Table~\ref{table:rmse_pcpnet} reports the RMSE of oriented normal predictions across different noise levels and point density variations. Our method achieves the lowest error under the highest noise level (1.2\%), indicating strong robustness to heavy corruption. It also performs competitively at moderate noise levels and under varying densities, ranking third overall in average RMSE behind NeuralGF and SHS-Net, the second of which is a fully supervised method. Notably, our approach outperforms several supervised baselines such as PCPNet and AdaFit, and consistently surpasses all classical methods by a significant margin. These results highlight our method’s ability to generalize well across challenging scenarios, despite not relying on supervised training signals.


\begin{table}[htbp]
\centering
\small
\caption{RMSE of oriented normals on PCPNet dataset. Our method achieves competitive performance even when compared to supervised baselines.}
\label{table:rmse_pcpnet}
\renewcommand\theadfont{\bfseries}
\setlength{\tabcolsep}{3pt}
\begin{tabularx}{\linewidth}{@{}l|cccc|cc|Y@{}}
\toprule
\multirow{2}{*}{\textsc{\textbf{Method}}} & \multicolumn{4}{c|}{\textsc{\thead{Noise Level}}} & \multicolumn{2}{c|}{\textsc{\thead{Density}}} & \multirow{2}{*}{\textsc{\thead{Avg}}} \\
 & None & 0.12\% & 0.6\% & 1.2\% & Stripe & Grad. & \\
\midrule
PCA + MST & 19.05 & 30.20 & 31.76 & 39.64 & 27.11 & 23.38 & 28.52 \\
PCA + SNO & 18.55 & 21.61 & 30.94 & 39.54 & 23.00 & 25.46 & 26.52 \\
PCA + ODP & 28.96 & 25.86 & 34.91 & 51.52 & 28.70 & 23.00 & 32.16 \\
LRR + MST & 43.48 & 47.58 & 38.58 & 44.08 & 48.45 & 46.77 & 44.82 \\
LRR + SNO & 44.87 & 43.45 & 33.46 & 45.40 & 46.96 & 37.73 & 41.98 \\
LRR + ODP & 28.65 & 25.83 & 36.11 & 53.89 & 26.41 & 23.72 & 32.44 \\

\hline
AdaFit + MST & 27.67 & 43.69 & 48.83 & 54.39 & 36.18 & 40.46 & 41.87 \\
AdaFit + SNO & 26.41 & 24.17 & 40.31 & 48.76 & 27.74 & 31.56 & 33.16 \\
AdaFit + ODP & 26.37 & 24.86 & 35.44 & 51.88 & 26.45 & 20.57 & 30.93 \\
HSurf ~~+ MST & 29.82 & 44.49 & 50.47 & 55.47 & 40.54 & 43.15 & 43.99 \\
HSurf ~~+ SNO & 30.34 & 32.34 & 44.08 & 51.71 & 33.46 & 40.49 & 38.74 \\
HSurf ~~+ ODP & 26.91 & 24.85 & 35.87 & 51.75 & 26.91 & 20.16 & 31.07 \\
PCPNet & 33.34 & 34.22 & 40.54 & 44.46 & 37.95 & 35.44 & 37.66 \\
SHS-Net & \bestcell{10.28} & \bestcell{13.23} & \bestcell{25.40} & 35.51 & \secondbestcell{16.40} & \secondbestcell{17.92} & \secondbestcell{19.79}\\
NeuralGF & \secondbestcell{10.60} & 18.30 & \secondbestcell{24.76} & \secondbestcell{33.45} & \bestcell{12.27} & \bestcell{12.85} & \bestcell{18.70} \\

\hline
\textbf{Ours} & {15.41} & \secondbestcell{17.98} & {25.70} & \bestcell{31.04} & 19.27 & {20.58} & 21.67 \\
\bottomrule
\end{tabularx}
\end{table}

\subsection{Further Ablation Studies}

To evaluate the impact of dictionary size and the cross-attention mechanism described in Section \ref{ssec:self_prior_field}, we conduct a series of controlled ablation experiments.\\

\begin{figure}[ht]
    \centering
    \includegraphics[width=\linewidth]{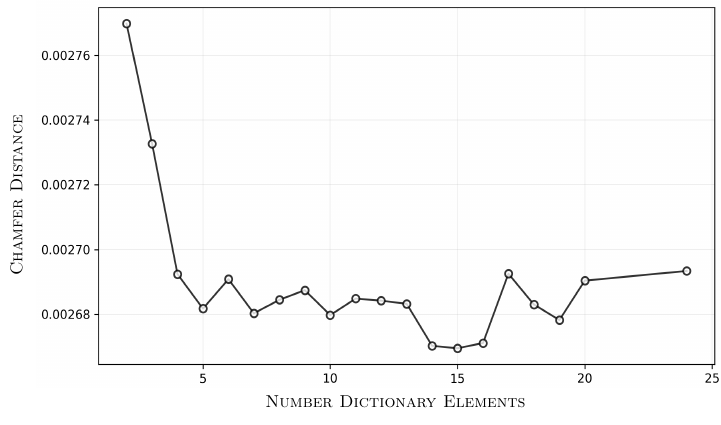}
    \caption{Plot shows the change in Chamfer Distance of the Neural-fields zero level-set against number of elements used within the dictionary.}
    \label{fig:ablationplot}
\end{figure}

We perform our analysis on the virus model from the self-similar dataset, where we expect local structure to benefit from increased dictionary expressiveness. We vary the dictionary size across a range of values from 2 to 20 and measure reconstruction quality using the Chamfer Distance between the predicted distance field and the ground truth. Specifically, we sample the predicted implicit surface defined by the attentive signed distance function (SDF), convert it to a point cloud, and compute the distance to the ground-truth point cloud. Results are plotted in Fig.~\ref{fig:ablationplot}.\\


We observe a clear trend: when the dictionary size is small (e.g., 2–4), the reconstruction is degraded. This is because the small dictionary primarily encodes coarse, global patterns, limiting expressiveness. As the dictionary size increases, reconstructions become progressively sharper and more faithful to the ground truth, indicating improved local pattern representation through richer token diversity. However, beyond a certain size, performance begins to saturate - this is accompanied by increasing similarity between tokens in the dictionary, suggesting redundancy. Based on this trade-off, we select a dictionary size of 16 for all main experiments, balancing accuracy and efficiency.\\


\section{Experimental Details}\label{experimental_details}

\subsection{Implementation and Environment}
All experiments were conducted using {PyTorch} with {PyTorch3D} for geometric operations. Training is performed on a single {NVIDIA RTX A5000 GPU}, with each shape taking approximately {8--12 minutes} to converge over {20{,}000 epochs}.

\subsection{Network Architecture}
We adopt an MLP architecture similar to that used in NeuralGF. The neural field $f_\theta$ is modeled using an 8-layer MLP with hidden dimension 256 and a skip connection at the midpoint layer. We apply {geometric initialization} as described in \cite{atzmon2020sald}, stabilizing the signed distance function near the zero level set.\\

To encode query coordinates, we use a sinusoidal {positional encoder} with 6 frequency bands. The encoded query is passed through a cross-attention module that interacts with a shared latent dictionary of geometric tokens. We use 8 attention heads in our multi-headed attention setup.

\subsection{Cross-Attention Prior}
The latent self-prior is implemented as a learnable {embedding dictionary} containing 16 tokens, initialized using QR decomposition of random matrices and updated via backpropagation. Cross-attention is applied between the encoded queries and dictionary tokens using multi-head attention, dynamically aggregating non-local geometric information across the shape.

\subsection{Training Procedure}
The training loss combines several self-supervised geometric terms, as detailed in the main paper; we use the following hyperparameters across all our experiments: $\alpha = 0.3$, $\beta = 10$, $\gamma = 1$, and $\delta = 0.01$. Training samples include both on-surface points from the input point cloud and off-surface points obtained via Gaussian perturbation. We follow the procedure introduced in \cite{li2023neuralgf}; to generate training samples, we first normalize the input mesh and downsample to a maximum of 300{,}000 points. For each point, we compute the distance to its 50th nearest neighbor and use this value as a local scale parameter. We then generate noisy query points by applying Gaussian perturbation scaled by the local distance and a global factor (\texttt{dis\_scale} = 0.15). For each query point, we identify its 64 nearest neighbors to construct local patches for geometric supervision. We generate up to 10 rounds of query points per shape, yielding a large, dense set of perturbed inputs and associated neighborhoods.\\

We train each shape independently using the {Adam} optimizer. We use a two-stage learning rate schedule: an initial {linear warm-up} phase followed by {cosine annealing}. During the first 10{,}000 iterations, the learning rate increases linearly from zero to the base learning rate of $1 \times 10^{-4}$. After the warm-up, the learning rate follows a cosine decay schedule until the end of training at 20{,}000 iterations. This approach encourages stable early training and smooth convergence.



\newcommand{\R}{\mathbb{R}} 
\newcommand{\norm}[1]{\left\lVert#1\right\rVert} 
\newcommand{\abs}[1]{\left\lvert#1\right\rvert} 
\newcommand{\set}[1]{\left\{#1\right\}} 
\newcommand{\GT}{\text{GT}}
\newcommand{\REC}{\text{REC}}

\section{Mesh Reconstruction Quality Metrics}
To quantitatively evaluate the quality of the reconstructed 3D meshes ($M_{\REC}$) against their corresponding ground truth meshes ($M_{\GT}$), we employ a suite of established geometric metrics. For metrics requiring point cloud representations, we uniformly sample $N_s$ points from the surfaces of both $M_{\GT}$ and $M_{\REC}$. Unless otherwise specified, $N_s = 100,000$ for Chamfer and Hausdorff distances, and $N_s = 10,000$ for F-Score computation.

\subsection{Chamfer Distance (CD)}
The Chamfer Distance measures the average squared distance between closest point pairs across two point sets. Let $S_{\GT} = \{ \bm{p}_1, \dots, \bm{p}_{N_s} \}$ be the set of points sampled from $M_{\GT}$, and $S_{\REC} = \{ \bm{q}_1, \dots, \bm{q}_{N_s} \}$ be the set of points sampled from $M_{\REC}$. The Chamfer Distance is defined as:
\begin{equation}
\begin{aligned}
d_{CD}(S_{\GT}, S_{\REC}) &= \frac{1}{|S_{\GT}|} \sum_{\bm{p} \in S_{\GT}} \min_{\bm{q} \in S_{\REC}} \|\bm{p} - \bm{q}\|_2^2 \\
&\quad + \frac{1}{|S_{\REC}|} \sum_{\bm{q} \in S_{\REC}} \min_{\bm{p} \in S_{\GT}} \|\bm{q} - \bm{p}\|_2^2
\end{aligned}
\end{equation}
where $\norm{\cdot}_2$ denotes the Euclidean L2-norm. A lower CD value indicates a better alignment between the two point sets, signifying higher reconstruction accuracy in terms of average surface proximity.

\subsection{Hausdorff Distance (HD)}
The Hausdorff Distance captures the maximum discrepancy between two point sets. It is a more stringent metric than CD as it is sensitive to outliers or localized large errors. Using the same point sets $S_{\GT}$ and $S_{\REC}$ as defined for CD, the Hausdorff Distance is given by:

\begin{equation}
\begin{aligned}
    d_{HD}(S_{\mathrm{GT}}, S_{\mathrm{REC}}) 
    &= \max\Bigg\{ 
        \sup_{\bm{p} \in S_{\mathrm{GT}}} \inf_{\bm{q} \in S_{\mathrm{REC}}} \|\bm{p} - \bm{q}\|, \\
    &\hphantom{=\max\Bigg\{} 
        \sup_{\bm{q} \in S_{\mathrm{REC}}} \inf_{\bm{p} \in S_{\mathrm{GT}}} \|\bm{q} - \bm{p}\| 
       \Bigg\}
\end{aligned}
\end{equation}
where $\sup$ denotes the supremum (least upper bound) and $\inf$ denotes the infimum (greatest lower bound). A lower HD value signifies a smaller maximum error between the surfaces.

\subsection{F-Score ($F_1$)}
The F-Score evaluates surface reconstruction quality by considering both precision and recall with respect to a distance threshold $\tau$. Points $P_{\GT}$ are sampled from $M_{\GT}$ and $P_{\REC}$ from $M_{\REC}$ (with $N_s=10,000$ samples for this metric).
Precision ($P$) is the fraction of points in $P_{\REC}$ that are within distance $\tau$ of any point in $P_{\GT}$:
\begin{equation}
    P(\tau) = \frac{1}{|P_{\REC}|} \sum_{\bm{q} \in P_{\REC}} \mathbb{I} \left( \min_{\bm{p} \in P_{\GT}} \norm{\bm{q} - \bm{p}}_2 < \tau \right)
\end{equation}
Recall ($R$) is the fraction of points in $P_{\GT}$ that are within distance $\tau$ of any point in $P_{\REC}$:
\begin{equation}
    R(\tau) = \frac{1}{|P_{\GT}|} \sum_{\bm{p} \in P_{\GT}} \mathbb{I} \left( \min_{\bm{q} \in P_{\REC}} \norm{\bm{p} - \bm{q}}_2 < \tau \right)
\end{equation}
where $\mathbb{I}(\cdot)$ is the indicator function, returning 1 if the condition is true, and 0 otherwise.
The F-Score is the harmonic mean of precision and recall:
\begin{equation}
    F_1(\tau) = 2 \cdot \frac{P(\tau) \cdot R(\tau)}{P(\tau) + R(\tau)}
\end{equation}
A higher F-Score (closer to 1) indicates better overall agreement between the surfaces, considering both completeness (recall) and correctness (precision).

\subsection{Normal Consistency (NC)}
Normal Consistency measures the alignment of surface normals between the reconstructed mesh $M_{\REC}$ and the ground truth mesh $M_{\GT}$. This metric is crucial for assessing the smoothness and geometric detail preservation of the reconstructed surface.
Let $F_{\REC}$ be the set of faces in $M_{\REC}$. For each face $f_i \in F_{\REC}$, let $\bm{c}_i$ be its centroid and $\hat{\bm{n}}_i$ be its unit normal vector.
We find the corresponding face $f_j^* \in F_{\GT}$ (the set of faces in $M_{\GT}$) whose centroid $\bm{c}_j^*$ is closest to $\bm{c}_i$:
\begin{equation}
    \bm{c}_j^* = \arg\min_{\bm{c}_k \in C_{\GT}} \norm{\bm{c}_i - \bm{c}_k}_2
\end{equation}
where $C_{\GT}$ is the set of all face centroids in $M_{\GT}$. Let $\hat{\bm{n}}_j^*$ be the unit normal of this closest ground truth face $f_j^*$.
The Normal Consistency is then computed as the average of the absolute dot products of these corresponding normal pairs:
\begin{equation}
    NC = \frac{1}{|F_{\REC}|} \sum_{f_i \in F_{\REC}} \abs{\hat{\bm{n}}_i \cdot \hat{\bm{n}}_j^*}
\end{equation}
The NC score ranges from 0 to 1, where 1 indicates perfect alignment of normals between the reconstructed mesh and the corresponding parts of the ground truth mesh. A higher NC score suggests that the reconstructed surface accurately captures the local orientation of the ground truth surface.

\subsection{Normal Estimation Metric}\label{appendix:normal_estimation}
The Oriented Root Mean Squared Error (RMSE\textsubscript{O}) quantifies the angular deviation between estimated surface normals and ground truth normals, taking orientation into account. This metric is crucial in applications where the direction of normals affects downstream tasks such as rendering or shading.  
Let $\hat{\bm{n}}_i$ and $\bm{n}_i$ denote the unit ground-truth and predicted normals, respectively, for each of the $I$ evaluation points. RMSE\textsubscript{O} is computed as:

\begin{equation}
    \text{RMSE}_O = \sqrt{\frac{1}{I} \sum_{i=1}^{I} \left( \arccos(\hat{\bm{n}}_i \cdot \bm{n}_i) \right)^2}
\end{equation}

The angular error is measured in degrees, ranging from $0^{\circ}$ (perfect alignment) to $180^{\circ}$ (opposite orientation). A lower RMSE\textsubscript{O} indicates more accurate normal orientation estimation, highlighting the fidelity of the reconstruction process.

\section{Qualitative Results Point2Mesh}
\centering
\begin{minipage}{\textwidth} 
    \centering
    \includegraphics[width=\linewidth]{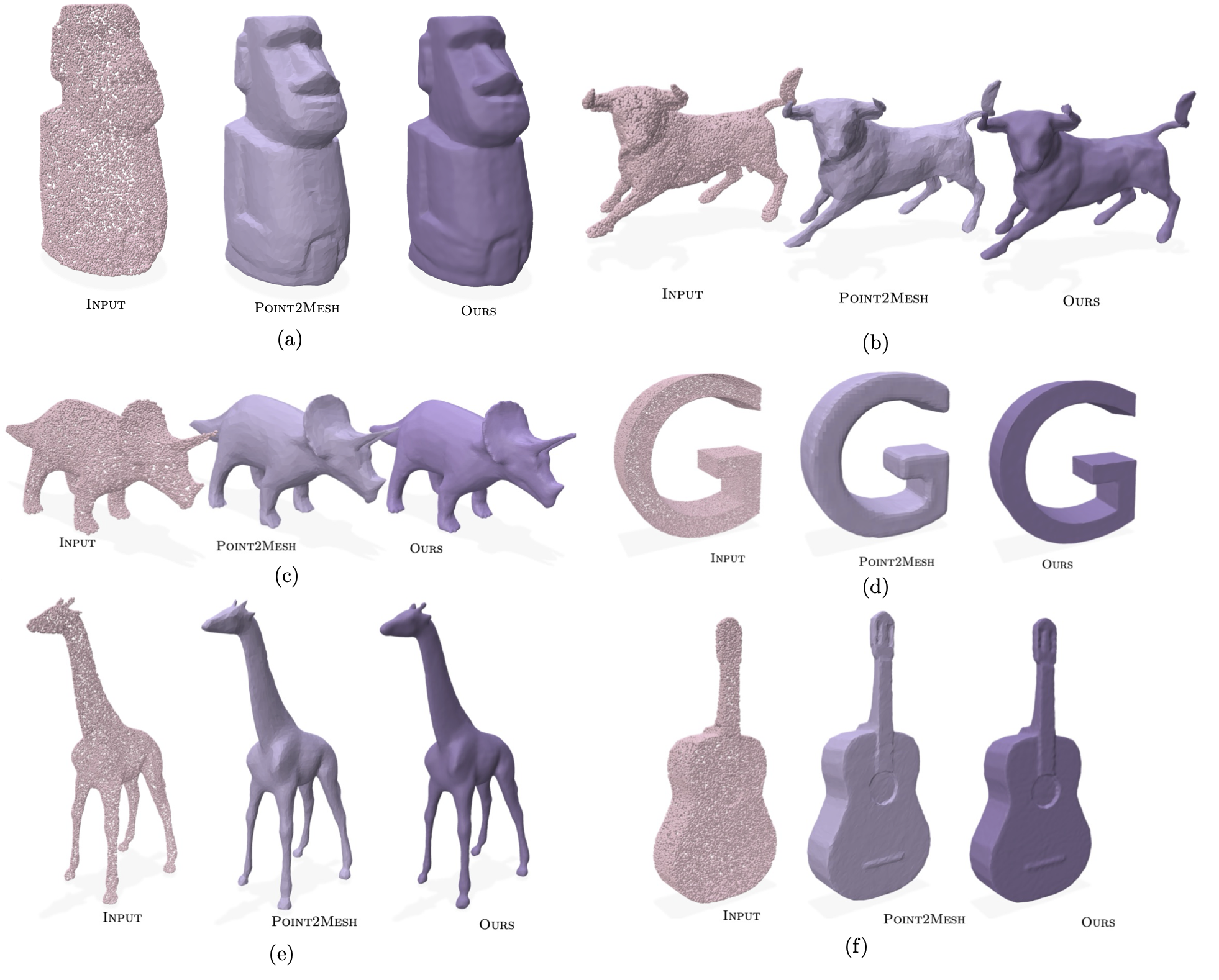}
    \captionof{figure}{We compare our approach with Point2Mesh \cite{hanocka2020point2mesh} using the publicly available objects released by the Point2Mesh authors. We note that in general our approach produces surfaces which are smoother while retaining sharp features.}
    \label{fig:p2m_dataset}
\end{minipage}

\newpage
\clearpage

\section{Qualitative Results on SRB}\label{QualitativeSRB}

\centering
\begin{minipage}{\textwidth} 
    \centering
    \includegraphics[width=\linewidth]{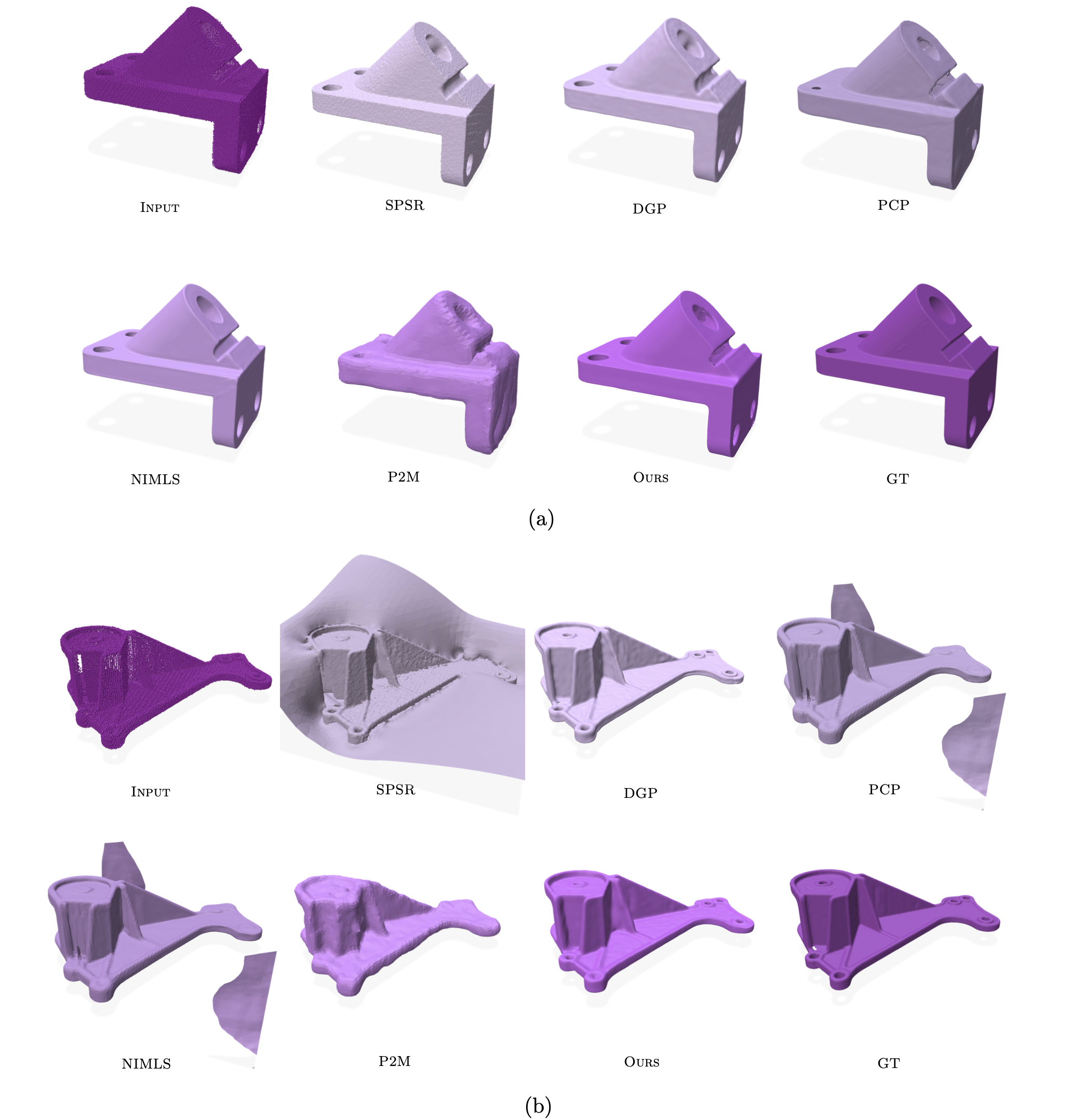}
    \captionof{figure}{Shows the qualitative results of our method on objects from the \textit{surface reconstruction benchmark} (SRB), compared against other reconstruction techniques. Methods are defined in Section \ref{sec:reconstruction}.}
    \label{fig:sbr_qual_1}  
\end{minipage}

\begin{figure*}
      \centering
    \includegraphics[width=\linewidth]{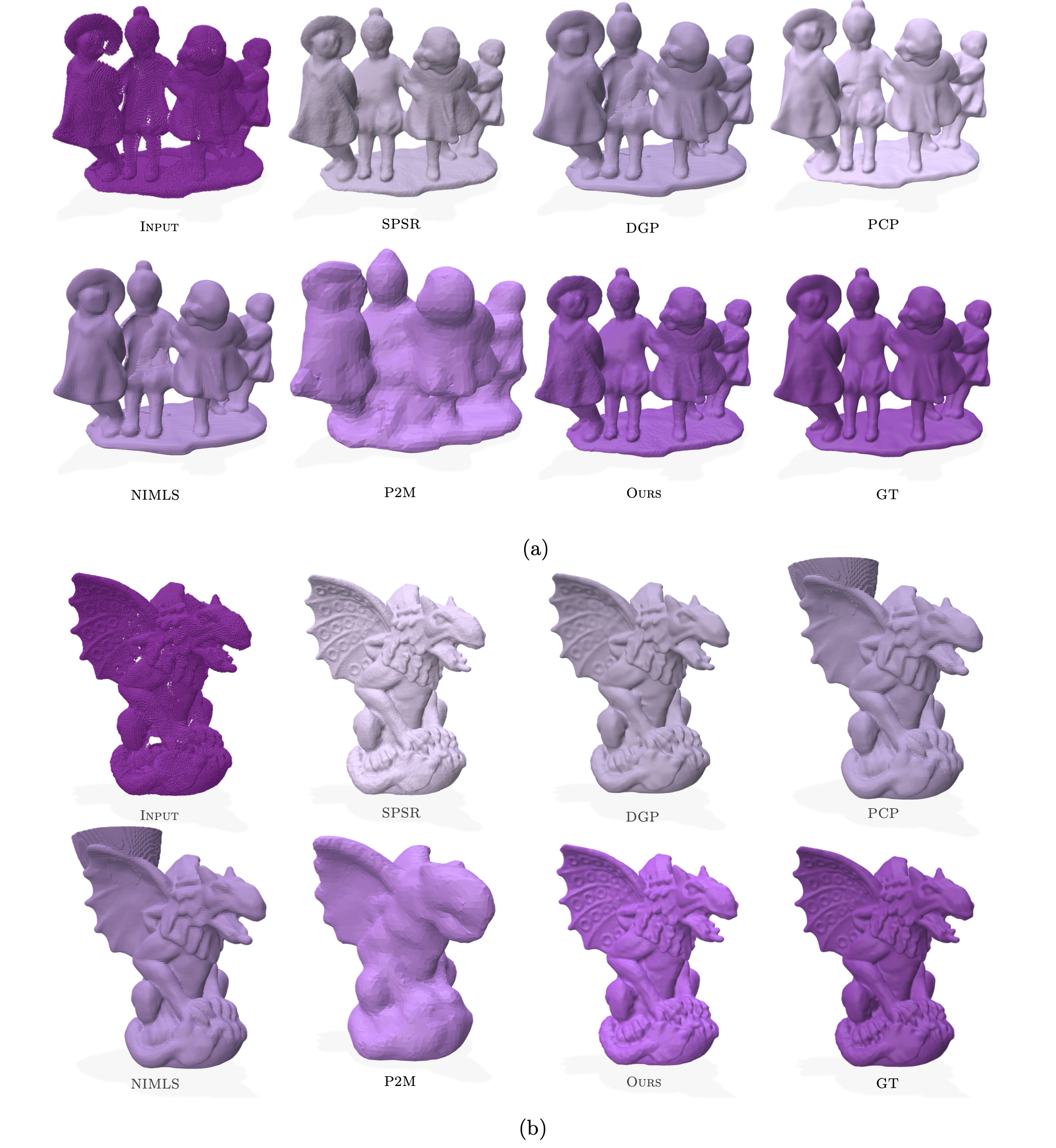}
    \caption{Shows the qualitative results of our method on objects from the \textit{surface reconstruction benchmark} (SRB), compared against other reconstruction techniques. Methods are defined in Section \ref{sec:reconstruction}.}
    \label{fig:sbr_qual_2}  
\end{figure*}
\begin{figure*}
      \centering
    \includegraphics[width=\linewidth]{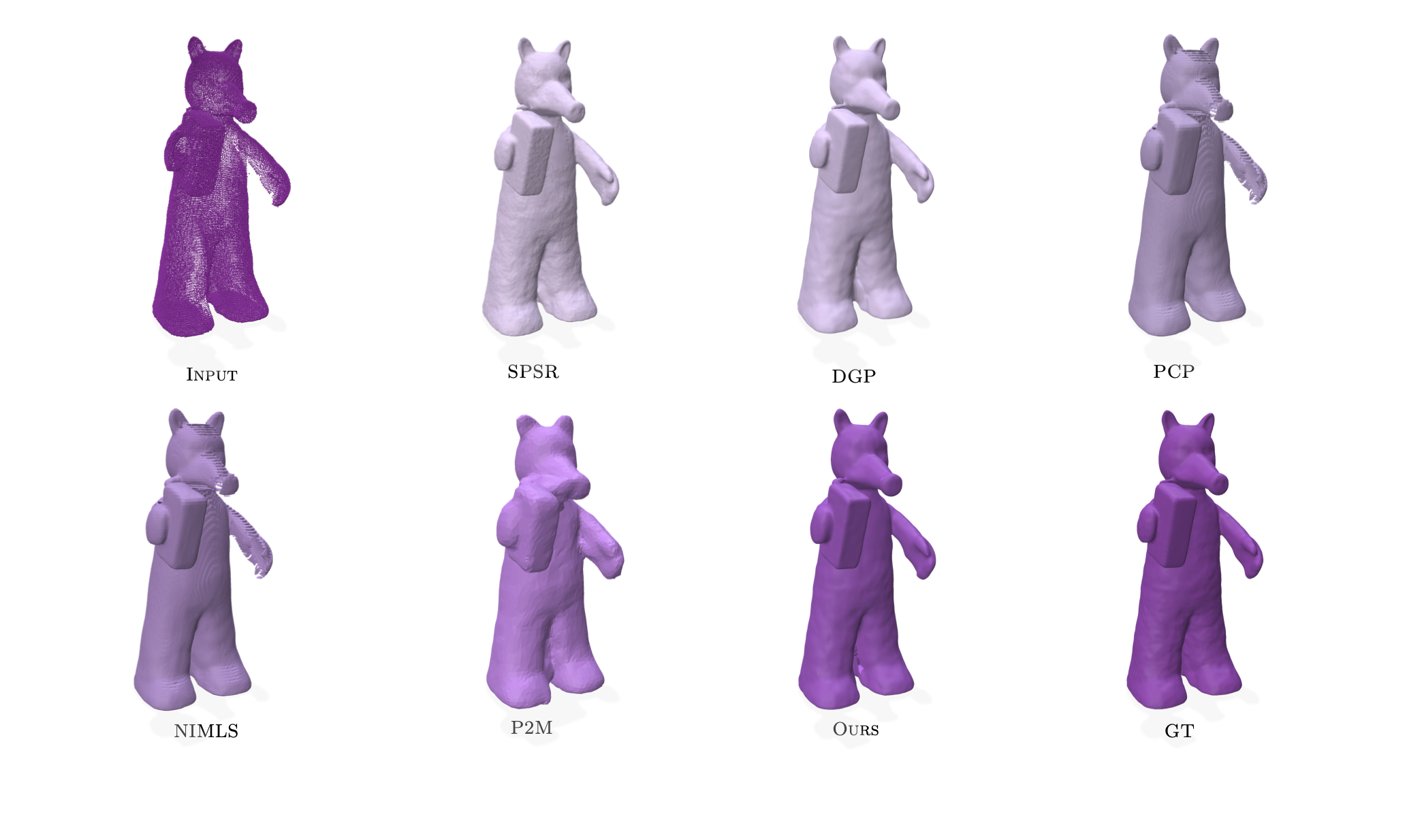}
    \caption{Shows the qualitative results of our method on objects from the \textit{surface reconstruction benchmark} (SRB), compared against other reconstruction techniques. Methods are defined in Section \ref{sec:reconstruction}.}
    \label{fig:sbr_qual_3}  
\end{figure*}
\newpage

\clearpage

\end{document}